\documentclass[sigconf]{acmart}

\copyrightyear{2026}
\acmYear{2026}
\setcopyright{cc}
\setcctype{by}
\acmConference[SIGSPATIAL '26]{The 34th ACM International Conference on Advances in Geographic Information Systems}{November 03--06, 2026}{Riverside, CA, USA}
\acmBooktitle{The 34th ACM International Conference on Advances in Geographic Information Systems (SIGSPATIAL '26), November 03--06, 2026, Riverside, CA, USA}
\acmDOI{10.1145/3841645.3843393}
\acmISBN{979-8-4007-2950-8/2026/11}

\begin{document}

\title{Non-Parametric Spatiotemporal Trajectory Prediction via State-Conditioned Transition Sampling}

\author{Michael Fore}
\correspondingauthor  
\affiliation{%
  \institution{Amazon Web Services}
  \city{Herndon}
  \state{VA}
  \country{USA}
}
\email{mikefore@amazon.com}

\author{Akshay Jain}
\affiliation{%
  \institution{Amazon Web Services}
  \city{Boston}
  \state{MA}
  \country{USA}
}
\email{jainaaks@amazon.com}

\author{Justin Downes}
\affiliation{%
  \institution{Amazon Web Services}
  \city{Arlington}
  \state{VA}
  \country{USA}
}
\email{jusdow@amazon.com}

\author{Rohan Pradhan}
\affiliation{%
  \institution{Amazon Web Services}
  \city{New York}
  \state{NY}
  \country{USA}
}
\email{awsrohan@amazon.com}

\author{Duncan Botti}
\affiliation{%
  \institution{Amazon Web Services}
  \city{Herndon}
  \state{VA}
  \country{USA}
}
\email{dunbotti@amazon.com}

\begin{abstract}
We present a training-free method for multi-modal trajectory prediction that achieves comparable accuracy to a 57M-parameter transformer while requiring no GPU and zero learned parameters. The method builds a transition table of historical state-to-next-position pairs and retrieves neighbors using a product kernel over spatial proximity, bearing, speed, and temporal context. Two inference modes operate over this shared representation: diversity-penalized sampling produces trajectories covering distinct plausible routes, while beam search finds the highest-likelihood path. On the TrAISformer benchmark (Danish Maritime AIS), our method achieves competitive accuracy at full data availability and dramatically outperforms the transformer in data-scarce regimes---remaining stable down to 10\% of training data where TrAISformer degrades catastrophically. This enables deployment in new geographic regions from an order of magnitude less historical data.
\end{abstract}

\ccsdesc[500]{Information systems~Geographic information systems}
\ccsdesc[500]{Mathematics of computing~Nonparametric statistics}
\ccsdesc[300]{Computing methodologies~Kernel methods}
\ccsdesc[300]{Applied computing~Transportation}

\keywords{trajectory prediction, nonparametric methods, kernel density estimation, maritime AIS, multi-modal prediction}

\maketitle

\section{Introduction}

Predicting future positions of moving entities is a core spatial computing problem with applications in maritime surveillance, traffic management, and search-and-rescue~\cite{liang2022vessel}. The task is multi-modal: at route junctions, objects may proceed in any of several directions and useful predictors must represent this uncertainty.

Deep learning approaches frame trajectory prediction as sequence modeling. TrAISformer~\cite{nguyen2024traisformer} discretizes AIS positions into grid cells and applies a transformer architecture to achieve state-of-the-art results on maritime trajectory prediction benchmarks. However, such models require GPU training, degrade with limited training data, and must be retrained for new geographic regions.

Nonparametric methods avoid these costs. Hexeberg et al.~\cite{hexeberg2017ais} proposed Single Point Neighbor Search (SPNS): retrieve the historical observations near the current position, discard those whose course deviates beyond a threshold, and dead-reckon on the median course and speed of the rest. This requires no training but produces a unimodal output and offers no uncertainty. Its successor NCDM~\cite{dalsnes2018ncdm} samples a course from the same gated set instead of collapsing it, giving several trajectories; both were built for collision avoidance at < 15 minute horizons. Other statistical approaches~\cite{ristic2008statistical,pallotta2013vessel} similarly rely on pattern extraction but lack multi-modal output.

Our approach bridges this gap by retrieving neighbors (state-to-next-position pairs) weighted by a product kernel over spatial proximity, bearing, speed, and time-of-day/day-of-week. We introduce two inference modes over this shared representation. The first, diverse sampling, applies a spatial repulsion penalty that forces successive trajectory samples to explore different routes; the second, beam search, maintains multiple candidate paths and prunes by cumulative path likelihood, producing single-trajectory predictions.

On the TrAISformer benchmark (DMA Danish maritime AIS, 3-hour prediction horizon), our method achieves a top-1 FDE of 9.00~km vs.\ TrAISformer's 9.84~km, and a best-of-16 FDE of 2.49~km vs.\ 2.66~km while requiring no GPU, no learned parameters, and only seconds of CPU setup. Particularly significant for deployment in new geographic regions: the method maintains its accuracy with as little as 10\% of training data, a regime where the transformer degrades catastrophically. Our contributions are:
\begin{enumerate}
    \item A state-conditioned transition predictor that replaces the hard course gate of neighbor search~\cite{hexeberg2017ais} with product-kernel weighting over spatial proximity, bearing, speed, and temporal conditioning.
    \item Diversity-promoting sequential sampling~\cite{vijayakumar2018diverse} applied to spatial transition lookup, producing multi-modal trajectory predictions that cover distinct plausible routes.
    \item Evaluation demonstrating that this training-free approach matches a 57M-parameter transformer at full data and dramatically outperforms it in data-scarce regimes, retaining its full-data accuracy on an order of magnitude less data.
\end{enumerate}

\section{Method}

Our method operates in two phases.\footnote{AI coding assistants were used to implement the method \& experiment scripts. The method design, experimental methodology, and interpretation of results are the authors'. All code was reviewed and validated by the authors.} First, we construct a transition table with spatial indices in an offline setup step. Second, we leverage the transition table in online inference via one of two modes: diverse sampling or beam search.

\subsection{Transition Table}

From training trajectories sampled at regular intervals (10 minutes in our experiments), we extract all pairs of observations separated by a fixed prediction step $\Delta t$ (1 hour). Each row stores the current state $\mathbf{s}_i = (\text{lat}_i, \text{lon}_i, v_i, \theta_i, h_i, d_i)$---position, speed, bearing, hour-of-day, and day-of-week---along with the next position $\mathbf{p}_i^{+} = (\text{lat}_i^{+}, \text{lon}_i^{+})$ observed $\Delta t$ later. The table is a flat collection of such records (150,000 rows, eight columns, about 9.6~MB), rather than a matrix over a discretized state space: nothing is binned, and the kernel of Section~\ref{sec:lookup} rather than a grid cell decides how much each row contributes. We build a BallTree spatial index on the current positions for neighbor retrieval.

\subsection{State-Conditioned Weighted Lookup}
\label{sec:lookup}
Given a query state $\mathbf{q} = (\text{lat}, \text{lon}, v, \theta, h, d)$ representing position, speed (knots), bearing (degrees), hour-of-day, and day-of-week, we retrieve all transitions within the spatial kernel's support and assign each transition $i$ a weight that is the product of five kernel terms: $w_i = K_{\text{sp}} \cdot K_{\text{brg}} \cdot K_{\text{spd}} \cdot K_{\text{hr}} \cdot K_{\text{dow}}$; where:
\begin{itemize}
    \item $K_{\text{sp}}$: Quartic kernel with adaptive bandwidth on great-circle distance, following Abramson's square-root rule~\cite{abramson1982bandwidth}.
    \item $K_{\text{brg}} = f_{\text{vM}}(\theta - \theta_i;\, \kappa)$: von Mises density on bearing difference.
    \item $K_{\text{spd}} = \exp(-{(v - v_i)^2}/{2\sigma_v^2})$: Gaussian on speed difference.
    \item $K_{\text{hr}} = \exp(-{\Delta_{\text{circ}}(h, h_i;\, 24)^2}/{2\sigma_h^2})$: cyclic Gaussian on hour-of-day.
    \item $K_{\text{dow}} = \exp(-{\Delta_{\text{circ}}(d, d_i;\, 7)^2}/{2\sigma_d^2})$: cyclic Gaussian on day-of-week.
\end{itemize}
Here $\Delta_{\text{circ}}(a, b;\, P) = \min(|a - b|,\; P - |a - b|)$ is the circular distance with period $P$.\footnote{Circular distance ensures that, for example, using a 24-hour period, hour 23 and hour 1 are two hours apart rather than twenty-two.} We omit month-of-year because the benchmark data cannot support it: ten of its twelve months are represented by only two or three days each, too sparse to estimate a seasonal effect. The result is a weighted set of next-positions $\{(\mathbf{p}_i^{+}, w_i)\}$ representing where entities in similar states went next. This is a form of Nadaraya-Watson~\cite{nadaraya1964,watson1964} kernel regression applied to state-conditioned spatial transitions.

\subsection{Autoregressive Inference}

Both inference modes generate trajectory branches autoregressively. At each step $t$ of a branch, the shared procedure is:
\begin{enumerate}
    \item Compute weights $\{w_i\}$ via the product kernel (Section~\ref{sec:lookup}).
    \item Optionally modify weights (mode-specific; see below).
    \item Sample one or more candidate next-positions from the weight distribution.
    \item Smooth each candidate: compute the weighted mean of all next-positions, modulated by a Gaussian ($\sigma\!=\!15^{\circ}$) on angular difference from the candidate's bearing. This mitigates single-sample noise that would otherwise compound across steps.
    \item Update the query state (position, bearing, hour-of-day) from the smoothed prediction and proceed to step $t+1$.
\end{enumerate}
The prediction horizon is $H$ steps of $\Delta t$ each. Speed is held fixed at the initial observed value for simplicity. Branches are ranked by cumulative unpenalized log-density.

\subsection{Diverse Sampling}

To produce $N$ branches that collectively cover plausible future routes, we take inspiration from Diverse Beam Search~\cite{vijayakumar2018diverse} by sampling branches sequentially with a spatial repulsion penalty applied to the observations in continuous space (as opposed to the discrete space version in the original paper). For branch $j > 1$, step (2) applies a spatial repulsion penalty:

\begin{equation}
w_i' = w_i \cdot \prod_{k=1}^{j-1} \left(1 - \exp\!\left(-\frac{\|\mathbf{p}_i^{+} - \mathbf{p}_k^{(t)}\|^2}{2\sigma_{\text{div}}^2}\right)\right)
\end{equation}
where $\mathbf{p}_k^{(t)}$ is the position of previously generated branch $k$ at step $t$. This suppresses observations whose next-positions are near existing branches, forcing successive branches to explore different routes. Each branch samples one candidate per step (step 3).

\subsection{Beam Search}

Beam search maintains $B$ active branches in parallel. Step (2) is skipped (no diversity penalty). At step (3), $C$ candidates are sampled per active branch, yielding $B \times C$ total candidates; only the top $B$ by cumulative log-density are retained. Paths on high-traffic corridors accumulate high scores; paths that wander off-route are pruned.

\begin{figure}[!t]
\centering
\includegraphics[width=\columnwidth]{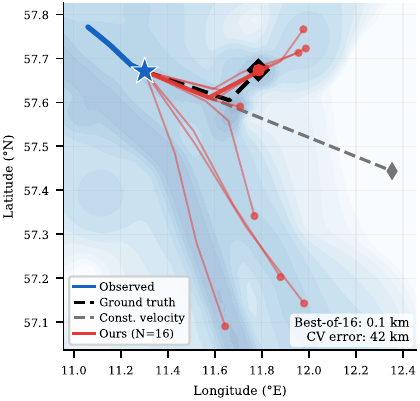}
\Description{Map showing diverse trajectory predictions at a maritime route junction, with 16 colored branches covering multiple shipping corridors.}
\caption{Diverse predictions covering multiple plausible shipping corridors at a route junction in the DMA dataset. Background shading shows traffic density (KDE, for visual context only---not used by the method). The best-of-16 branch lands within 0.5~km of ground truth.}
\label{fig:junction}
\end{figure}

\section{Experiments}

\subsection{Setup}

\textbf{Dataset.} We evaluate on the DMA AIS dataset~\cite{dma}, covering vessel traffic in the Danish Straits region, sampled at 10-minute intervals. We follow the split from ~\cite{nguyen2024traisformer}: 1,481 test trajectories (March 21--31, 2019) and 10,605 training trajectories (2,517 vessels from 2019 excluding March 21--31). Each trajectory has a 3-hour observation window followed by a 3-hour prediction horizon.

\textbf{Baselines.}
\begin{itemize}
\setlength{\itemsep}{0pt}
\setlength{\parskip}{0pt}
    \item \emph{Constant velocity (CV)}: Linear extrapolation from the last observed position and bearing.
    \item \emph{Hexeberg SPNS}~\cite{hexeberg2017ais}: Dead reckoning on the median course and speed of neighbors within a radius, gated at $35^\circ$ of course deviation, applied autoregressively. Produces one trajectory (no multi-modal output).
    \item \emph{NCDM}~\cite{dalsnes2018ncdm}: SPNS's multi-modal successor, sampling a course per branch from the same gated neighbors instead of taking the median, giving $N\!=\!16$ unranked trajectories. \footnote{Our implementation of both neighbor-search baselines advances a fixed time step rather than the published fixed distance, and draws neighbors from our spatial index rather than a 100~m radius.}
    \item \emph{Spatial-only + diversity}: Ablation of our method using spatial kernel weighting only (no bearing, speed, or temporal conditioning), with diversity penalty ($N\!=\!16$, $\sigma_{\text{div}}\!=\!5$~km). Isolates the contribution of multi-dimensional conditioning.
    \item \emph{LSTM seq2seq}: 2-layer encoder-decoder LSTM (401,924 parameters), trained for 50 epochs with MSE loss. Best-of-16 via Gaussian noise injection on the hidden state.
    \item \emph{TrAISformer}~\cite{nguyen2024traisformer}: 57.4M-parameter transformer trained for 50 epochs. Top-1 via argmax decoding; best-of-16 via categorical sampling.
\end{itemize}

\noindent While recent surveys~\cite{liang2022vessel} cover various deep learning approaches to maritime trajectory prediction, we compare against TrAISformer as it is the only published method with both public code and multi-modal evaluation on the DMA benchmark. We report results from our own reproduction.

\textbf{Metrics.}
We report Final and Average Displacement Error (FDE, ADE) in kilometers at 1-hour and 3-hour horizons---the haversine distance between predicted and true position at the horizon endpoint, and averaged over the path up to it. We report \emph{top-1} (error of the single best-ranked prediction) and, following standard multi-modal evaluation practice, \emph{best-of-$N$} (minimum error across $N$ predictions), with $N=16$ as in TrAISformer. Since our predictions are hourly and the baselines' are every ten minutes, ADE is computed with our waypoints linearly interpolated onto the finer grid. To assess probabilistic calibration, we additionally report the Continuous Ranked Probability Score (CRPS) in its energy form: $\text{CRPS} = \mathbb{E}[\|X - y\|] - \frac{1}{2}\mathbb{E}[\|X - X'\|]$, where $X, X'$ are samples and $y$ is the ground truth.

\textbf{Parameters.} We cap our transition table at 150,000 transitions subsampled from training data and select the following hyperparameters based on data characteristics (e.g., typical route separation, AIS measurement precision). Kernel bandwidths: $\kappa\!=\!16$ (bearing, effective within ${\approx}\pm 20^{\circ}$), $\sigma_v\!=\!1$~knot (speed), $\sigma_h\!=\!6$h (hour-of-day), $\sigma_d\!=\!2$d (day-of-week). Beam search: $B\!=\!16$, $C\!=\!3$. Diverse sampling: $N\!=\!16$, $\sigma_{\text{div}}\!=\!5$~km. These are held fixed without systematic tuning. A formal sensitivity analysis is left to future work.

\subsection{Results}
\label{sec:main_results}

\textbf{Main results.} Table~\ref{tab:main} presents the comparison on the DMA test set. Beam search achieves the best 3-hour top-1 FDE and the best top-1 ADE, while diverse sampling achieves the best 3-hour best-of-16 FDE. TrAISformer retains the best 1-hour top-1 and the best 3-hour best-of-16 ADE, though neither ADE difference is significant under a paired bootstrap. On CRPS both learned baselines outperform both of our modes.

\begin{table*}[t]
\caption{Trajectory prediction on DMA AIS (1,481 test trajectories, 3-hour prediction). FDE and ADE in km. Single-route baselines have no best-of-16; NCDM's samples are exchangeable, so it has no top-1.}
\label{tab:main}
\small
\begin{tabular}{lrrrrrrrrrr}
\toprule
& & \multicolumn{2}{c}{1h top-1} & \multicolumn{2}{c}{1h best-16}
& \multicolumn{2}{c}{3h top-1} & \multicolumn{2}{c}{3h best-16} & \\
\cmidrule(lr){3-4} \cmidrule(lr){5-6} \cmidrule(lr){7-8} \cmidrule(lr){9-10}
Method & Params & FDE & ADE & FDE & ADE & FDE & ADE & FDE & ADE & 3h CRPS \\
\midrule
Const.\ velocity & 0 & 3.67 & 1.79 & -- & -- & 19.31 & 8.17 & -- & -- & -- \\
Hexeberg SPNS~\cite{hexeberg2017ais} & 0 & 5.74 & 3.14 & -- & -- & 18.51 & 9.32 & -- & -- & -- \\
NCDM~\cite{dalsnes2018ncdm} & 0 & -- & -- & 4.13 & 2.41 & -- & -- & 11.76 & 6.54 & 15.15 \\
Spatial-only + div. & 0 & 14.92 & 8.84 & 2.72 & 1.83 & 42.00 & 22.96 & 8.90 & 5.20 & 32.52 \\
LSTM seq2seq & 402K & 3.12 & 2.70 & 1.62 & 2.11 & 9.05 & 4.86 & 4.78 & 3.58 & 7.84 \\
TrAISformer & 57.4M & \textbf{2.20} & \textbf{1.32} & \textbf{0.81} & \textbf{0.70} & 9.84 & 4.42 & 2.66 & \textbf{1.76} & \textbf{6.02} \\
\midrule
Ours (beam) & 0 & 2.29 & 1.49 & 1.99 & 1.33 & \textbf{9.00} & \textbf{4.31} & 7.17 & 3.57 & 8.29 \\
Ours (diverse) & 0 & 4.29 & 2.66 & 1.01 & 0.86 & 14.10 & 7.19 & \textbf{2.49} & 1.85 & 9.00 \\
\bottomrule
\end{tabular}
\end{table*}

\begin{figure}[!t]
\centering
\includegraphics[width=\columnwidth]{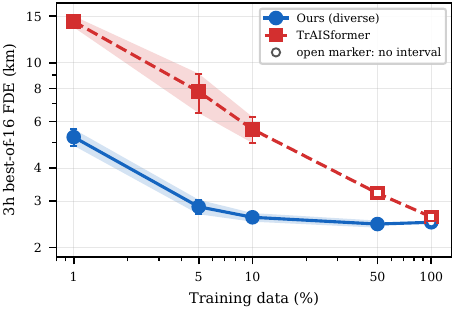}
\Description{Line plot on logarithmic axes of best-of-16 FDE versus training data fraction, with shaded 95 percent confidence bands. Our method stays near 2.5 km from 100 percent down to 10 percent of the data before rising, while TrAISformer rises steadily from 2.6 km at full data to 14.3 km at 1 percent. Our band is several times narrower than TrAISformer's throughout.}
\caption{Data efficiency: 3-hour best-of-16 FDE vs.\ training data fraction, with 95\% bands over subset draws. Log axes; open markers have no interval.}
\label{fig:sparsity}
\end{figure}

\textbf{Data efficiency.} To evaluate how both methods degrade as training data is reduced, we train on subsets of voyages from January--November 2019 and test on December (290 trajectories), a different split from the main results (in order to vary training set size while maintaining a held-out evaluation). Full data is 10,266 voyages carrying observations on 67 days---nearer ten weeks of regional traffic than the eleven months it spans---so the 10\% setting is roughly 1 week. For our method only, the transition table is capped at 150k transitions, which binds above 22\% of the pool, so 50\% and 100\% build identical tables. Figure \ref{fig:sparsity} illustrates that our method's performance is stable from 100\% down to 10\% of training data (2.46--2.60~km). Degradation becomes measurable at 5\% and substantial at 1\% (5.24~km). TrAISformer degrades from 2.61~km at full data to 14.33~km at 1\%---a 5.5$\times$ increase. At 10\% of data our method is 2.2$\times$ more accurate than TrAISformer.

\textbf{Computational requirements.} On an AMD EPYC 9R14, inference costs 46~ms per trajectory for diverse sampling and 14~ms for beam search; building the transition table takes 21~s at full data, falling to 0.3~s at 1\%. The cap bounds index and per-query cost but not extraction. Training TrAISformer took 69 minutes on a T4 GPU.

\section{Discussion and Conclusion}
Our method is effective across a broad operating range: it matches learned models at full data availability while offering dramatically better performance in data-scarce scenarios.

Beam search's best-of-16 (7.17~km) is weaker than diverse sampling (2.49~km), as expected---without a diversity mechanism its branches cluster on the same high-likelihood route. The spatial-only ablation's top-1 (42.00~km) is worse than SPNS (18.51~km) for an analogous reason: diversity sampling optimizes for multi-modal coverage at the expense of top-1, whereas SPNS deterministically follows its neighbors' median course. Adding bearing, speed, and temporal conditioning reduces best-of-16 to 2.49~km, a 72\% improvement over spatial-only, quantifying the value of multi-dimensional kernel conditioning; NCDM, which samples the same neighbor set uniformly, reaches only 11.76~km. Diverse sampling's CRPS (9.00~km) is higher than beam's (8.29~km) because the diversity penalty deliberately places samples on routes the vessel did not take, increasing expected distance to ground truth---the same property that yields its best-of-16 advantage. Both trail learned baselines, so we claim accuracy and data efficiency rather than calibration.

TrAISformer retains an advantage at the 1-hour top-1 metric (2.20 vs.\ 2.29~km), where its attention over the 3-hour observation window helps disambiguate routes at junctions. At the 3-hour horizon, TrAISformer generates 16 branches by independently sampling from its categorical distribution at each step; since most probability mass concentrates in the dominant direction, repeated samples cluster on the same route. Argmax decoding selects the highest-probability bin independently at each attribute (lat, lon, SOG, COG), which can produce positions between modes at junctions. Beam search scores complete paths by cumulative log-density, naturally favoring trajectories that stay on shipping lanes throughout.

\textbf{Limitations and outlook.} Without physics-based conditioning, our method cannot represent behavior absent from the transition table, such as very rare maneuvers. The fixed-speed assumption limits accuracy for vessels that accelerate mid-trajectory. No single inference mode dominates all metrics, requiring practitioners to select the mode appropriate to their use case. In very dense traffic, the neighbor count per query grows; the table cap bounds this in practice, but scaling in high-density regions warrants further study. 

Deployment in a new region requires only accumulating AIS data, and the transition table can be updated incrementally as it arrives. Finally, our evaluation covers a single region and a single entity type; the method's nature suggests applicability to other domains (pedestrian traffic, aviation, etc.) but we leave this to future work.

\bibliographystyle{ACM-Reference-Format}
\bibliography{references}

\end{document}